\documentclass[10pt,twocolumn,letterpaper]{article}

\usepackage[pagenumbers]{cvpr} 

\usepackage[dvipsnames]{xcolor}

\usepackage{dsfont} 
\usepackage{mathtools} 
\usepackage[mathscr]{euscript}
\newcommand\sbullet[1][.5]{\mathbin{\vcenter{\hbox{\scalebox{#1}{$\bullet$}}}}}
\usepackage{amsmath}
\usepackage{amssymb}
\DeclareMathOperator*{\argmin}{arg\,min}
\newcommand\norm[1]{\lVert#1\rVert}

\usepackage{graphicx}
\usepackage{booktabs}
\usepackage{multirow} 
\usepackage{subcaption} 
\usepackage{color, colortbl,xcolor}
\definecolor{Gray}{gray}{0.9}
\definecolor{light-gray}{gray}{0.95}

\usepackage{url} 

\usepackage{microtype}
\usepackage{pifont}
\newcommand{\cmark}{\textcolor{green!80!black}{\ding{51}}}
\newcommand{\xmark}{\textcolor{red}{\ding{55}}}
\date{} 

\definecolor{cvprblue}{rgb}{0.21,0.49,0.74}
\usepackage[pagebackref,breaklinks,colorlinks,citecolor=cvprblue]{hyperref}

\title{Synergy and Synchrony in Couple Dances} 

\author{
    Vongani Maluleke \textsuperscript{1} \quad
   Lea M{\"u}ller\textsuperscript{1} \quad
    Jathushan Rajasegaran \textsuperscript{1} \quad
  Georgios Pavlakos\textsuperscript{2} \quad \\
  Shiry Ginosar\textsuperscript{1} \quad
  Angjoo Kanazawa\textsuperscript{1} \quad
   Jitendra Malik \textsuperscript{1}\\
   \\
   \textsuperscript{1}UC Berkeley 
   \textsuperscript{2}UT Austin
}

\begin{document}

\twocolumn[{
    \renewcommand\twocolumn[1][]{#1}
    \maketitle
    \vspace{-2.6em}
    \centering
    \includegraphics[width=1.0\linewidth,trim={0cm 3cm 0 6cm},clip]{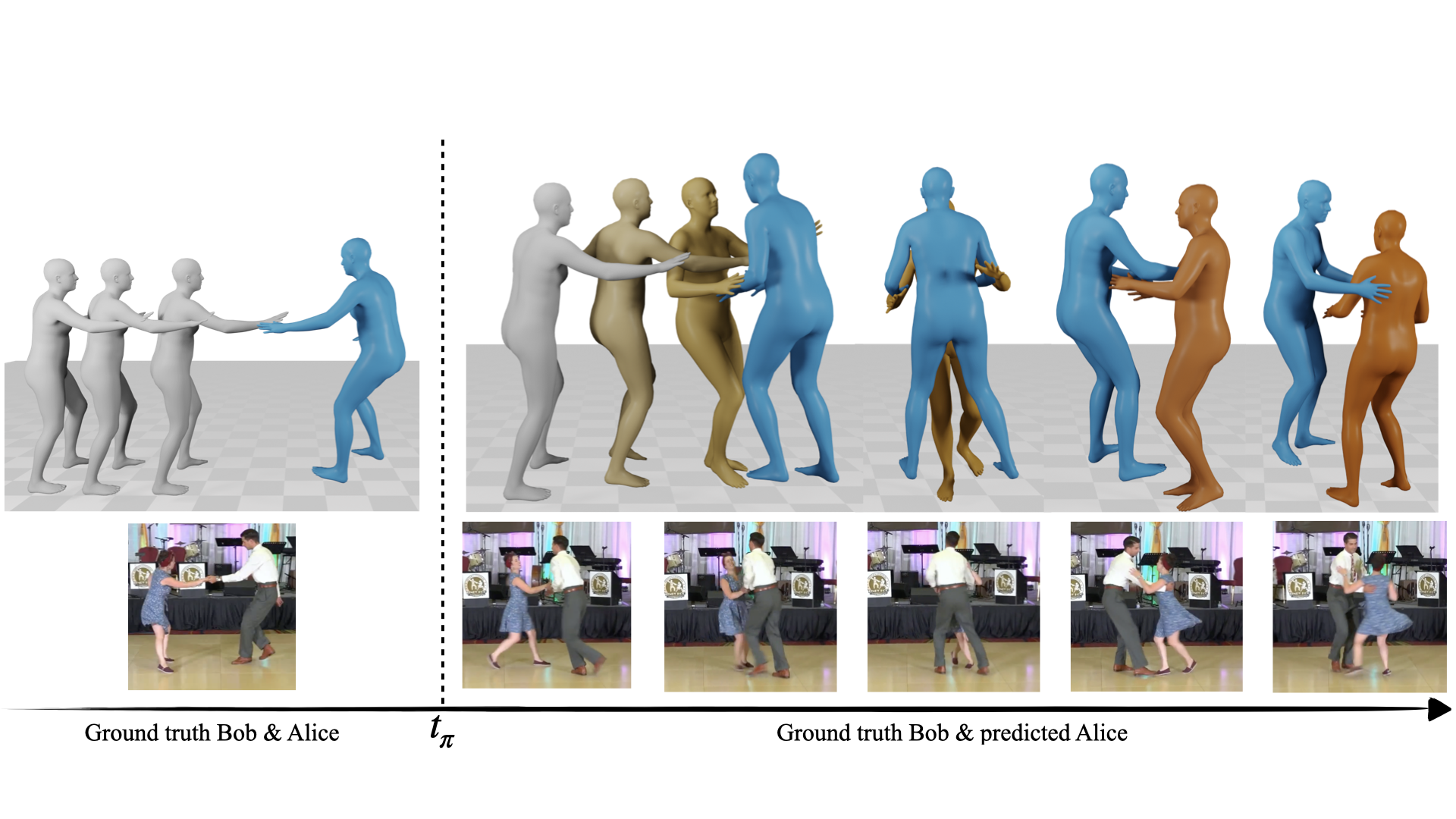}
    \captionof{figure}{
    \textbf{To what extent does Bob's behavior affect Alice's behavior?} We study this question in a couple's dance - an example of full-body dyadic physical social interaction. We predict the full body motion of a dancer, Alice (orange), given their own past motion (gray) and their partner, Bob's (blue), motion.
    }
    \label{fig:teaser}
    \vspace{1.0cm}
}]

\begin{abstract}
This paper asks to what extent social interaction influences one's behavior. We study this in the setting of two dancers dancing as a couple. We first consider a baseline in which we predict a dancer's future moves conditioned only on their past motion without regard to their partner. We then investigate the advantage of taking social information into account by conditioning also on the motion of their dancing partner. We focus our analysis on Swing, a dance genre with tight physical coupling for which we present an in-the-wild video dataset. We demonstrate that single-person future motion prediction in this context is challenging. Instead, we observe that prediction greatly benefits from considering the interaction partners' behavior, resulting in surprisingly compelling couple dance synthesis results (see supp. video). Our contributions are a demonstration of the advantages of socially conditioned future motion prediction and an in-the-wild, couple dance video dataset to enable future research in this direction. 
Video results are available on the project website: \url{https://von31.github.io/synNsync}.
  
\end{abstract}    
\section{Introduction}

What is the counterpart in video understanding of next-word prediction, the bedrock pre-training task of large language models? Is a ``word'' in video a pixel? Or a patch? Or an entity like an object or a person? While these are debatable questions, this paper will focus on a human as the token of interest, with an associated ``state'', the counterpart of the word embedding vector used in natural language processing.

Unlike next-token prediction in language, the dynamics of a person's state are conditioned on much more than the past state history of the person, as we also have to consider their interactions with other people during social situations. For example, when one partner raises their arm during a couple dance, the other partner is likely to twirl.

This paper investigates how social information from interaction partners impacts human future behavior prediction.
We consider this question in the context of physically close social interactions such as couple dance, where we predict the future motion of one of the dancing partners. In social situations, single-person future behavior prediction is possible to some extent, thanks to the \textit{synergy} created by the coupling body parts resulting in a lower dimensional manifold of human motion than the available degrees of freedom~\cite{bernstein}. However, interacting partners dynamically and reciprocally adapt the temporal structure of their behavior. This \textit{synchrony}~\cite{riley2011interpersonal,knoblich2011psychological,delaherche2012interpersonal} enforces constraints on coordinated motion that are unpredictable from each partner alone and that we must take into account in social situations. We therefore examine how the dancers' prediction improves when considering their dancing partner's motion.

\looseness=-1 We represent continuous dance motion as a series of atomic motion elements. To understand why this might be useful, we can look at Labanotation~\cite{von1928schrifttanz}, a notation invented for capturing dance motion like a musical score. The idea behind Labanotation is to record the positions of the body, the pattern of the steps, and places of emphasis. Labanotation thus breaks dance down into atomic motions with associated discrete notations that are easier to write down on paper and analyze than the continuous motion because there is a set dictionary of them.

Similarly, we learn a discrete dictionary of quantized atomic motion elements with a motion-encoding VQ-VAE~\cite{van2017neural}. Critically, our motion representation relies on a parametric 3D human body model, a computationally crucial choice since it allows us to ignore all the irrelevant pixels and focus on predicting behavior directly. Moreover, the parametric body model enables us to decouple motion into its constituents: pose, orientation, and translation, and learn a separate atomic dictionary for each---a necessary detail because dancers move freely around the stage.

The synchrony between interacting partners hints that there is enough temporal correlation between their behaviors for predicting the motion of one agent based on that of the other -- but to what extent? We study this question with an autoregressive transformer~\cite{vaswani2017attention} as our prediction mechanism, as transformers excel at capturing long-range temporal correlations in time-series data. The discrete motion representation learned by the VQ-VAE is well-suited for transformers  and enables computationally efficient prediction using discrete classification rather than continuous regression. It also allows for nondeterministic prediction as the output of the autoregressive transformer is a multinomial distribution over the next timestep of motion, from which we can sample multiple trajectories.

We demonstrate the efficacy of leveraging social information in a set of experiments on a dataset of in-the-wild Swing couple dance videos, which we introduce and release as part of this work. Our results show that social information from an interaction partner greatly enhances the predictability of human behavior. Our contributions are a demonstration of the advantages of socially conditioned future motion prediction and an in-the-wild, annotated couple dance video dataset to enable future research in this direction.

\section{Related Work}

\begin{table}\centering \footnotesize
\begin{tabular}{@{}lccccc@{}}\toprule 
& 3D mesh & In-the-wild & Future pred. \\
\midrule
Qianhui \etal \cite{baruah2020multimodal}                       & \xmark & \xmark & \xmark \\
ReMoS \cite{ghosh2024remos}       & \xmark & \xmark & \xmark \\
Murchana \etal \cite{murchana2020multimodal} & \xmark & \xmark & \cmark \\
InterFormer \cite{baptiste2023interaction} & \xmark & \cmark & \xmark \\
Duolando \cite{siyao2024duolando}                & \cmark & \xmark & \xmark \\
ExPI \cite{guo2022multi} & \cmark & \xmark & \cmark \\

\rowcolor{Gray}
Ours                        & \cmark & \cmark & \cmark
 \\
\bottomrule
\end{tabular}
\caption{Related work on multi-person motion generation and estimation. We are the first to demonstrate how internet data can be leveraged to train a model for predicting 3D human motion capturing the full body surface.}
\label{tab:related}
\end{table}

\medskip
\noindent

\textbf{Human Motion Prediction} involves generating future human motion from past data. The well-studied case of \textit{unary human motion prediction} aims to predict a person's future motion form their own past motion. For this task, traditionally, Recurrent Neural Networks (RNNs) were used \cite{Fragkiadaki2015RecurrentNM, julieta2017motion}, while recent work leverages attention mechanisms \cite{vaswani2017attention, Cai2020LearningPJ}. Some studies predict unary motion in interactions with objects or scenes \cite{corona2020context,cao2020long,mao2022contact}, based on input pose \cite{starke2019neural,taheri2022goal,zhang2022couch}, or conditioned on scene contact \cite{rempe2021humor} or audio signals \cite{ng2022learning2listen}. Recently, Radosavovic \etal \cite{TokenHumanoid2024} used next token prediction for humanoid locomotion. 

\textit{Dyadic Human Motion Prediction} extends unary prediction by incorporating past motion from a second person \cite{guo2022multi,murchana2020multimodal}; please see \Cref{tab:related} where we refer to this task as ``Future pred.''. 

Existing methods for multi-person motion prediction are typically trained on data collected in controlled settings, using MoCap systems \cite{guo2022multi} or depth sensors \cite{murchana2020multimodal}. While in-lab data can produce clean motion sequences, it suffers from limitations in scalability and naturalness.
In contrast, we demonstrate that the problem of social interaction can be studied using real video data from the Internet, allowing us to analyze the motion of professional swing dancers through state-of-the-art 4D human reconstruction methods. Our approach in principle can be applied to any collection of Internet videos.

\medskip
\noindent
\textbf{Motion Generation} addresses the problem of synthesizing human motion from scratch. While many studies focus on generating single-person motions from various modalities, \eg, speech audio~\cite{ginosar2019learning,alexanderson2020style,liu2022beat,yi2023generating}, text~\cite{Ghosh_2021_ICCV,Guo_2022_CVPR,petrovich22temos,TEACH:3DV:2022,zhang2023generating,tevet2023human}, scene~\cite{hassan2021stochastic,wang2022towards,huang2023diffusion}, action~\cite{guo2020action2motion,petrovich2021action,tevet2022human}, music \cite{tang2018dance,aiozGdance,li2020learning,li2021learn,siyao2022bailando,tseng2023edge}, research on generating dyadic human motions is still emerging. A few prior works have explored dyadic human mesh generation during close social interactions \cite{shafir2023human,liang2023intergen,mueller2023buddi}. 
Closest to our approach, previous work has explored predicting a listener's 3D motion from a speaker's motion and audio in conversational settings \cite{jonell2019learning,ng2022learning2listen,zhou2022responsive}. Multiple works generate a person's motion conditioned on the full (past, current, and future) sequence of another person \cite{siyao2024duolando,baptiste2023interaction,baruah2020multimodal,ghosh2024remos}; we mark these methods with an ``\xmark'' in \Cref{tab:related} under ``Future pred.''. Unlike these methods, where the model has access the full conditioning signal, our goal is to predict two people's future motion only from their past motion.

\medskip
\noindent
\textbf{VQ-VAEs for Human Motion Representations.}
The idea of learning a discrete dictionary of atomic motion elements for the purpose of motion generation was first introduced by~\cite{ng2022learning2listen} for facial motion of stationary individuals captured by a static frontal camera and recently expanded to full-body motion~\cite{zhang2023generating,dwivedi_cvpr2024_tokenhmr}.

These previous works are trained to compress multiple motion parameters, \ie, 3D joint locations~\cite{zhang2023t2m,siyao2024duolando}, SMPL \cite{bogo2016keep} body pose parameters \cite{dwivedi_cvpr2024_tokenhmr}, and facial motion \cite{ng2022learning2listen}, into a single codebook.

While some body poses naturally correlate with a person's orientation, most poses can occur in any orientation or translation. A unified codebook entangles these elements, limiting its ability to capture diverse human poses particularly in the context of social interaction. In this work, we propose to learn three separate dictionaries for body pose, orientation, and translation parameters of a human body model for each person, and show the effectiveness of this approach compared to a single codebook through an ablation experiment.

\medskip
\noindent
\textbf{Datasets of People in Close Social Interaction.}
Various labs have captured data of two people interacting closely. Hi4D \cite{yin2023hi4d}, CHI3D \cite{fieraru2023reconstructing} contain social interactions between two people such as ``hugging'' or ``shaking hands'' with video and ground-truth body shaped from motion capture, respectively. InterGen \cite{liang2023intergen} is a dataset of interacting people captured via motion capture and annotated with text descriptions. ReMoS \cite{ghosh2023remos} is a motion capture dataset of people performing martial arts and couple dance, respectively. ExPi \cite{guo2022multi} contains sequences of two people in ``extreme pose interactions'' of professional dancers captured in a motion capture studio.

\section{Motion Prediction in Couple Dance}

Our aim is to investigate the practical benefits of incorporating social information into future behavior prediction during physical interaction. We focus on motion prediction as a proxy of behavior and predict the motion of a dancer engaged in couple dance. We study two aspects of motion during physical interaction: how a person's past motion predicts their future and how individuals modulate their motions when engaged in close, coordinated activities. For notation purposes, we introduce Alice and Bob, a pair of fictional characters commonly used as placeholders in cryptography literature~\cite{rivest1978method}, whom we assign to represent the two people engaged in the dance.
We refer to Alice and Bob's specific components via the superscripts A and B. \eg, $\textbf{M}^A$ refers to Alice's motion, and $\textbf{M}^B$ refers to Bob's motion.

We define the following two tasks: (1) Unary motion prediction: \emph{given Alice's past 3D-body motion, we autoregressively predict her future moves.}; and (2) Dyadic motion prediction: \emph{given Alice and Bob's past 3D-body motion, we autoregressively predict the future motion of Alice, who is dancing together with Bob.}

\begin{figure*}
    \centering
    \includegraphics[width=0.9\linewidth]{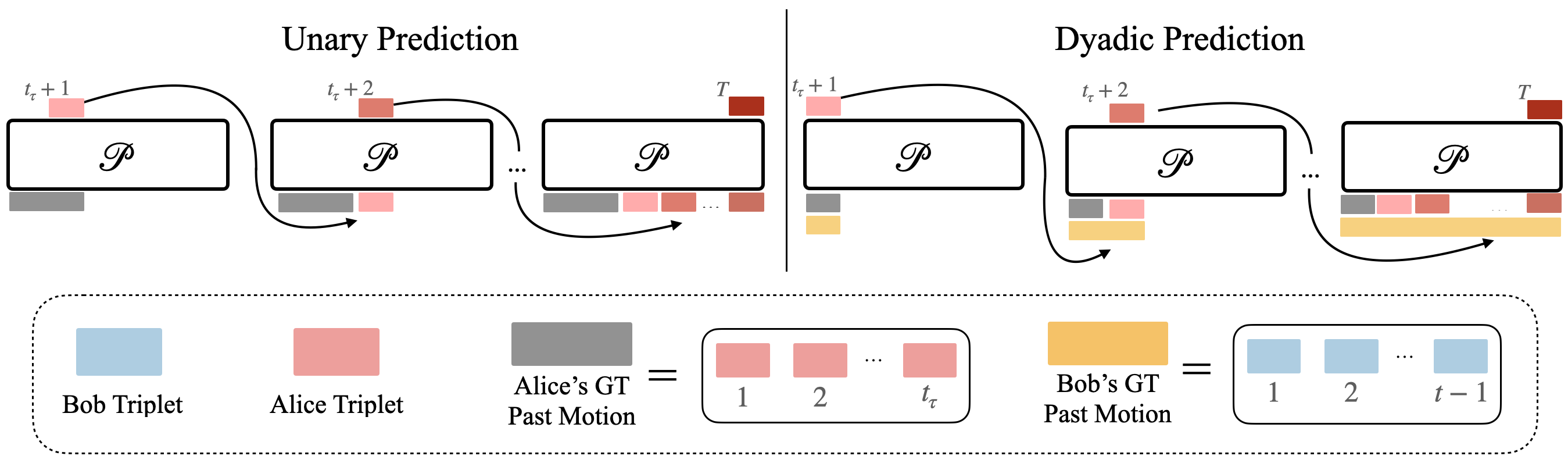}
    \vspace{-0.2cm}
    \captionof{figure}{Test-time autoregressive prediction in the unary (left) and dyadic (right) tasks. In the unary prediction task, we start from Alice's past motion up to time $t_\pi$, and predict the next time step in each iteration. In the dyadic prediction task, we start from Alice \textit{and} Bob's motion up to time $t_\pi$. In each step, we predict Alice's next token from her past ground truth motion until $t_\pi$, her predicted motion for $t > t_\pi$, and Bob's past ground truth motion}
    \vspace{-0.3cm}
    \label{fig:sampling}
\end{figure*}

To represent the flow of motion, we define a 
transformer-based 
predictor, $\mathcal{P}$, that learns to model temporally long-range patterns in the input sequence. At test time, the predictor takes as input either Alice's past motion (in the unary case) or Alice and Bob's past motion (in the dyadic case) and autoregressively predicts Alice's next dance steps (Figure~\ref{fig:sampling}).

We represent the continuous dance motion using a dictionary of atomic motion elements, that we learn using a set of motion VQ-VAEs (Section~\ref{sec:vqvae}). As dancers may freely move around space, we extend motion VQ-VAE modeling by disentangling the three full-body motion constituents: body pose, orientation and translation. The discrete representation learned by the VQ-VAEs enables us to predict a multinomial distribution over the next timestep of motion. Thus, the output of the autoregressive predictor  (Section~\ref{ss:transformer}) is a distribution over Alice's possible realistic future dance moves, from which we can sample multiple trajectories.

\subsection{Problem Definitions}

\medskip
\noindent
\textbf{Unary Motion Prediction.} 
Let $\mathbf{M}=\{\mathbf{m}_i\}_{i=1}^T$ be a pose trajectory representing motion $\mathbf{m}_i$. For each timestep $t \in [t_\pi,T]$, where $t_\pi \in [1,T]$ is the time in which we start predicting,
we take as input Alice's past ground truth motion up to $t_\pi$, $\mathbf{{M}}^A_{1:t_\pi}$ and any of her previously predicted past motion $\mathbf{\hat{M}}^A_{t_{\pi}+1:t-1}$,
if available.
Our predictor, $\mathcal{P}$, autoregressively predicts Alice's  future motion one time-step at a time:
\begin{equation}
       \mathbf{\hat{m}}^A_{t} = \mathcal{P}( \mathbf{{M}}^A_{1:t_\pi},\mathbf{\hat{M}}^A_{t_{\pi}+1:t-1}), \\
\end{equation}
where $\mathcal{P}$ learns to model the distribution over the next timestep of Alice's motion:
\begin{equation}
p(\mathbf{\hat{m}}^A_{t} | \mathbf{M}^A_{1:t-1}).
\end{equation}

\medskip
\noindent
\textbf{Dyadic Motion Prediction.} 
Let $\mathbf{M}=\{\mathbf{m}_i\}_{i=1}^T$ be a pose trajectory  representing motion $\mathbf{m}_i$. For each timestep $t \in [t_\pi,T]$, where $t_\pi \in [1,T]$ is the time in which we start predicting,
we take as input Bob's past ground truth motion
$\mathbf{M}^B_{1:t-1}$
along with Alice's past ground truth motion up to $t_\pi$, $\mathbf{{M}}^A_{1:t_\pi}$ and any of her previously predicted past motion $\mathbf{\hat{M}}^A_{t_{\pi}+1:t-1}$,
if available.
Our predictor, $\mathcal{P}$, autoregressively predicts Alice's future motion one time-step at a time:
\begin{equation}
       \mathbf{\hat{m}}^A_{t} = \mathcal{P}(\mathbf{M}^B_{1:t-1}, \mathbf{{M}}^A_{1:t_{\pi}},\mathbf{\hat{M}}^A_{t_\pi+1:t-1}), \\
\end{equation}
where $\mathcal{P}$ learns to model the distribution over the next timestep of Alice's motion, given Bob's motion:
\begin{equation}
p(\mathbf{\hat{m}}^A_{t} | \mathbf{M}^B_{1:t-1}, \mathbf{M}^A_{1:t-1}).
\end{equation}

\subsection{Learning Quantized Motion Codebooks} 
\label{sec:codebooks}

\medskip
\noindent
\textbf{Motion Representation.} An important question is how to represent the human body in the world. Ideally, the body representation enables us to generalize across individuals and camera viewpoints. Unlike previous methods that, among others, rely on 3D joint locations, we choose a representation that is invariant to changes in body shape, scale, and camera pose. In particular, we use the SMPL~\cite{bogo2016keep} body model to represent human bodies in 3D. SMPL is a differentiable function that maps pose, $\mathbf{\theta} \in \mathbb{R}^{J \times 3}$ and shape, $\mathbf{\beta} \in \mathbb{R}^{B}$ parameters to a 3D mesh with $J=23$ joints and $B<300$. Through $\mathbf{\phi} \in \mathbb{R}^{3}$ and $\mathbf{\gamma} \in \mathbb{R}^{3}$, the body can be oriented and translated in the 3D world. To represent motion over time $T$, we extend the parameters related to the pose trajectory via $\mathbf{\Theta} \in \mathbb{R}^{T \times J \times 3}$, $\mathbf{\Phi} \in \mathbb{R}^{T \times 3}$, and $\mathbf{\Gamma} \in \mathbb{R}^{T \times 3}$. We further denote a person's pose trajectory of duration $T$ via: 
$$\mathbf{P} = \{\mathbf{m}_i\}_{i=1}^T = \{[\mathbf{\theta}, \mathbf{\phi}, \mathbf{\gamma}]\}_{i=1}^{T} = [\mathbf{\Theta}, \mathbf{\Phi}, \mathbf{\Gamma}] \text{.}$$

\medskip
\noindent
\textbf{VQ-VAE's for Full-Body Motion.} 
\label{sec:vqvae}
VQ-VAE's~\cite{van2017neural} are encoder-decoder models trained to compress input data into a lower dimensional discrete latent space, \ie, ``the codebook''. The encoder maps input data into a continuous latent space, and a quantization process maps these continuous representations to discrete codes in the codebook. Given a sequence of codes, the decoder reconstructs the corresponding continuous input data.

The discrete codebook $\mathcal{Z} = \{z_k\}_{k=1}^{K}$ of a VQ-VAE consists of $K$ codebook entries $z_k \in \mathbb{R}^{d_z}$. Given an input signal $\textbf{X} \in \mathbb{R}^{T \times d_x}$, we encode $\textbf{X}$ via $\hat z = E(x)$ to a latent variable of size $\frac{T}{w} \times d_z$, where $w$ is a temporal window size usually much smaller than $T$. A quantization function $\textbf{q}(\cdot)$ then maps each element $\hat z_i$, $i=1,\dots,\frac{T}{w}$, to the index, $\bar z$, of its closest entry in $\mathcal{Z}$ via:

$$\bar z = \textbf{q}(\hat z) := \argmin_{z_k \in \mathcal{Z}}(\norm{\hat z_i - z_k}) \text{.}$$

Given the index $\bar z$ of a codebook entry, the decoder's task is to recover the original input signal: 
$$\hat x = D(\bar z) = D(\textbf{q}(E(x))) \text{.}$$

\begin{figure}
    \centering
    \includegraphics[width=1\linewidth]{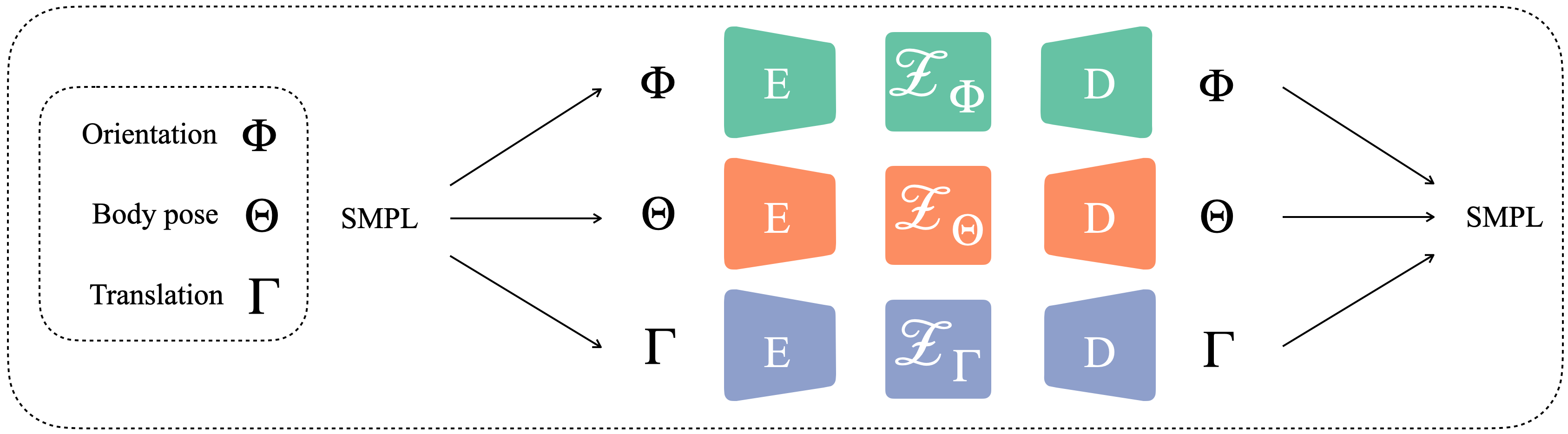}
    \caption{Illustration of the VQ-VAEs. We learn three separate codebooks, one for each body model parameter. An encoder, $E_{\cdot}$, maps the body parameter to the codebook, $\mathcal{Z}_{\cdot}$. The decoder, $D_{\cdot}$ brings the codebook latent vectors back into body model parameter space. To obtain 3D meshes, we jointly pass the parameters through the body model function.}
    \label{fig:vqvae}
\end{figure}

To model the full range of dance motion, we must consider all the constituents of motion: body pose, orientation, and translation. While these parameters may correlate in simple trajectories (\eg, walking forward or walking in a circle), in most cases, one can strike the same pose at any global orientation and translation. Therefore, we extend upon the commonly used unified codebook approach, which entangles the different aspects of motion~\cite{ng2022learning2listen,zhang2023t2m}. Instead, we factorize the dictionary learning and learn three separate codebooks, one for
each body model parameter (see~\cref{fig:vqvae}).

We refer to these codebooks via $\mathcal{Z}_{\Theta}$, $\mathcal{Z}_{\Phi}$, and $\mathcal{Z}_{\Gamma}$ for pose, orientation, and translation, respectively. 
The pose trajectory, or motion sequence $\mathbf{s}$, of a single person given in body model parameter space can then be approximated through a triplet of VQ-VAE codebook indices, by combining indices from each codebook, \ie,
$$\mathbf{\hat P} = \mathbf{s} = (\bar z_{\Theta}, \bar z_{\Phi}, \bar z_{\Gamma}) \text{.}$$

\subsection{Conditional Autoregressive Prediction}
\label{ss:transformer}

Our goal is to capture long-range temporal correlations in the input data to autoregressively predict Alice's future motion. To do this, we a train a transformer-decoder based network directly on the learned codebook indices.

We train the transformer with a cross entropy loss on the codebook indices,

\begin{equation*}
\label{eq:transformerloss}
    \mathscr{L_\mathcal{P}} = \mathds{E}_{y \sim p(y)}[-\log(p(\mathbf{s}^A_{\tau+1})] = 
    - \sum_{\tau=1}^{T} \sum_{k=1}^{K} \mathbf{s}^A_{t} \log(p(\mathbf{s}^A_{\tau+1}),
\end{equation*}
where the target codebook index at $\tau+1$ is computed from Alice's ground truth future motion
$y = \mathbf{M}^A_{t+1:t+1+w}$.

\medskip
\noindent
\textbf{Unary Autoregressive Prediction.} In the unary prediction scenario, the input sequence consists of the indices of Alice's body model triplets:

$$\mathbf{s}^A_{1:\tau} = (\bar z^A_{{\Theta}_{1:\tau}}, \bar z^A_{{\Phi}_{1:\tau}}, \bar z^A_{{\Gamma}_{1:\tau}}) \text{,}$$ 
where $\tau$ is the length of the encoded sequence.
The network's task is to predict the multinomial distribution of the codebook indices of the next token triplet:
 $$\mathcal{P}(\mathbf{s}^A_{1:\tau}) = \mathbf{s}^A_{\tau+1} = (\bar z^A_{{\Theta}_{\tau+1}}, \bar z^A_{{\Phi}_{\tau+1}}, \bar z^A_{{\Gamma}_{\tau+1}}) \text{.}$$

\medskip
\noindent
\textbf{Dyadic Autoregressive Prediction.} To compress Bob's motion, a straightforward way would be to re-use the VQ-VAE codebooks from the unary prediction task. This representation, however, does not model the interaction and relative motion between Alice and Bob. Instead, we train three new codebooks on Bob's body model parameters in Alice's reference frame. In particular, given Alice's body model parameters at time $t=0$, \ie, body orientation, $\mathbf{\Phi}_0^A$, and translation, $\mathbf{\Gamma}_0^A$, we model Bob's orientation, $\mathbf{\Phi}^B$, and translation, $\mathbf{\Gamma}^B$, \wrt Alice by transforming:
\begin{equation}
    \begin{aligned}
    \mathbf{\Gamma}^{B \text{wrt.} A} &=& (\mathbf{\Phi}_0^A)^{-1} (\mathbf{\Gamma}^B - \mathbf{\Gamma}_0^A) \\ 
    \mathbf{\Phi}^{B \text{wrt.} A} &=& (\mathbf{\Phi}_0^A)^{-1} (\mathbf{\Phi}^B)^{-1}  \text{.}
    \end{aligned}
\end{equation}
Then, we train $\mathcal{Z}_{\Phi}^B$ on $\mathbf{\Phi}^{B \text{wrt.} A}$ and $\mathcal{Z}_{\Gamma}^A$ on $\mathbf{\Gamma}^{B \text{wrt.} A}$. $\mathcal{Z}_{\Theta}^B$ is trained on $\mathbf{\Theta}^{B}$, and for Alice we use the codebooks from the unary task, $\mathcal{Z}_{\Theta}^A$, $\mathcal{Z}_{\Phi}^A$, and $\mathcal{Z}_{\Gamma}^A$. 

In the dyadic scenario, instead of only using Alice's motion as input to the transformer, we concatenate Bob's motion along the token dimension. Specifically, given Alice and Bob's motion sequences in VQ-VAE indices, $\mathbf{s}^{A} \in \{1, \dots, K\}^{(3 \times \frac{T}{w})}$ and $\mathbf{s}^{B} \in \{1, \dots, K\}^{(3 \times \frac{T}{w})}$, the predictor's input becomes:
$$\mathbf{s}^{AB} = [\mathbf{s}^A|\mathbf{s}^B] \in \{1, \dots, K\}^{(3 \times \frac{2T}{w})} \text{.}$$

Given initial motion sequences of Alice and Bob:
$$\mathbf{s}^A_{1:\tau} = (\bar z^A_{{\Theta}_{1:\tau}}, \bar z^A_{{\Phi}_{1:\tau}}, \bar z^A_{{\Gamma}_{1:\tau}}) 
\text{ and }
\mathbf{s}_{1:\tau}^B = (\bar z^B_{{\Theta}_{1:\tau}}, \bar z^B_{{\Phi}_{1:\tau}}, \bar z^B_{{\Gamma}_{1:\tau}})
\text{,}$$ 
the networks task is to predict the next token triplet for Alice:
$$\mathcal{P}(\mathbf{s}^A_{1:\tau} \text{ and } \mathbf{s}^B_{1:\tau}) = \mathbf{s}^A_{\tau+1} = (\bar z^A_{{\Theta}_{\tau+1}}, \bar z^A_{{\Phi}_{\tau+1}}, \bar z^A_{{\Gamma}_{\tau+1}}) \text{.}$$

\begin{figure}
    \centering
    \includegraphics[width=1\linewidth]{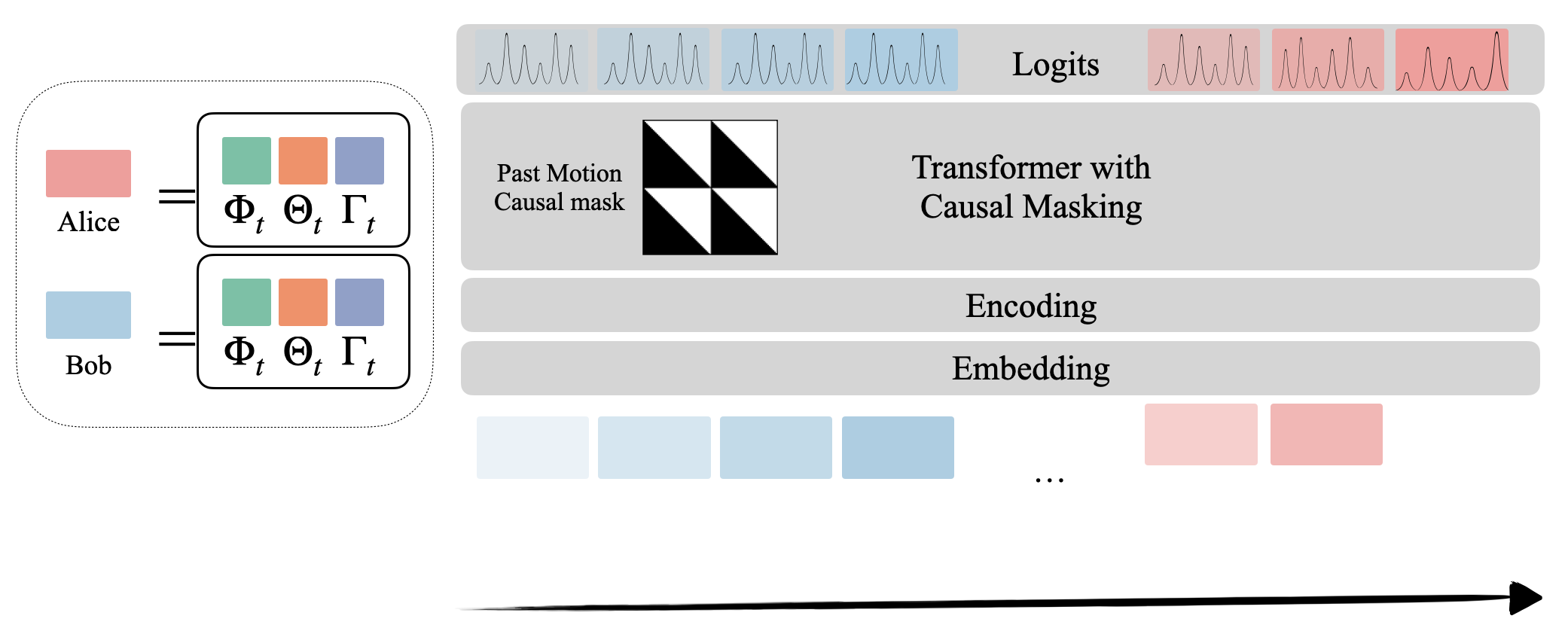}
    \caption{Transformer training procedure in the dyadic case. The major part of our network is a transformer-decoder block with causal masking, such that Alice and Bob can only attend to their past motion. Input to our model are Alice and Bob's codebook indices for body pose, $\Theta$, orientation, $\Phi$, and translation, $\Gamma$. We embed the tokens into a latent space and add time, person, and parameter encoding. The final layer in our network generates probability scores over codebook indices, representing the likelihood of an index being the next motion $\mathbf{s}_{t_{\pi}+1}$.}
    \label{fig:transformer}
\end{figure}

\medskip
\noindent
\textbf{Transformer Architecture.}
The predictor, $\mathcal{P}$, is designed with embedding and encoding layers, a transformer decoder block, and a final layer to predict logits (see \cref{fig:transformer}). Specifically, we first embed the input dictionary indices
$\mathbf{s} \in \mathbf{\{1, \dots K\}}^{(3 \times \frac{T}{w})}$ 
into d-dimensional latent variables $\mathbf{D} \in \mathbf{R}^{\frac{3T}{w} \times d}$ via an embedding layer. Note that we shift the codebook indices of $\bar z_{\Phi}$ by $K$ and those of  $\bar z_{\Gamma}$ by $2K$ such that indices of each codebook are distinct. Before passing $\mathbf{D}$ to the transformer decoder, we add time, person, and parameter encoding layers. We employ causal masking in the decoder such that each token can only attend to previous time steps.
Finally, the predictor $\mathcal{P}$ outputs 
$p(\mathbf{M}^A_{t}) \in \{\mathds{R}^{K}\}^3$,
\ie, the multinomial distributions of Alice's motion next codebook indices across each codebook's $K$ entries. 

\subsection{Implementation Details}
Our models are trained with a batch size of 128 and a learning rate of $1e-5$.

Our VQ-VAE codebooks have 1024 indices and 256-dimensional latent vectors. Each video sequence of length 100 frames is downsampled by a factor of $w=4$.
At test time, we use teacher forcing on Alice's input motion sequence, $\mathbf{M}_{1:\tau}$. To select codebook indices from predicted logits, we use top-k sampling with $k=103$. For all our experiments, we use 48 frames as input / query motion and predict the remaining 52 frames.

\section{3D Human Couple Dancing Dataset}
\label{sec:data}

We collected competitive Swing dance videos from the 2014-2021 International Lindy Hop Championship competitions (\url{https://ilhc.com/}) that were publicly distributed on YouTube. During this time frame, ILHC competitions were filmed with one mostly static frontal camera. We filtered the videos to only ones that contain couple dances (rather than a single dancer) and do not show the audience. We also removed introductory slides, including text related to the competition's or dancers' names. The final dataset contains 
$\mathbf{30}$ \textbf{hours} of clean couple dance footage. 
For the experiments in this paper, we segment each video into 4-second chunks.
The dataset will be publicly released.

To lift the motion of the couple to 3D, we use SLAHMR~\cite{ye2023decoupling}, which recovers the 3D motion of the couple in a world coordinate frame by considering and decoupling the motion of the camera~\cite{teed2021droid} from the motion of the people.
To achieve the best results, we use 4D Humans~\cite{goel2023humans} to track the people and estimate their 3D pose to initialize SLAHMR.
One challenge we faced here is related to estimating the height of the two partners, which is difficult from monocular observations.
Errors in relative height can significantly deteriorate the 3D output since the relative positions of the two people are incorrect.
To remedy that, we assume that one of the two partners will be slightly taller than the other (since we primarily have male-female couples).
This assumption introduced an extra constraint for the relative heights of the two people in the SLAHMR optimization (see Supp.), which improves the relative placement of the two dancers.

SLAHMR represents motion in an arbitrary world coordinate frame, similar to observing the motion as a member of the audience.
We canonicalize the motion sequences by introducing an extra transformation to make our analysis invariant to the observed viewpoint.
Effectively, the person of interest at time $t=0$ is at the origin of the coordinate frame, with a canonical orientation (\ie, the torso facing in the +$z$ direction and the head in the +$y$ direction).
Moreover, as a form of data augmentation, we mirror the recovered motions along the $yz$ plane.

\section{Evaluation}
We evaluate our motion prediction results on metrics designed to measure the quality of Alice's predicted motion in the unary and dyadic setups. In the dyadic case, we additionally evaluate the motion coordination between Alice and Bob as a couple. Since we use all available information from both interaction partners to predict the dyadic case, we further compare Alice's dyadic motion prediction to several strong baselines, as outlined below.

\medskip
\noindent \textbf{Evaluation Metrics.}
Quantifying motion realism is a difficult problem that cannot be reduced to a single metric. We thus evaluate our predictions along multiple axes following~\cite{ng2022learning2listen,ng2024audio2photoreal}. Our evaluation suite is based on the notion that humans should display (1) realistic and (2) diverse motion that is (3) synchronous with the motion of their interaction partner when they are interacting socially. We assess Alice's generated motion according to these three pillars against her raw ground truth motion $y$. 

    \noindent $\sbullet$ \textbf{FD}  (unary \& dyadic): Motion realism measured by the Frechet distance~\cite{heusel2017gans} between generated and ground truth motion distributions. We directly calculate the FD on the 3D joint locations of the predicted meshes.
    
    \noindent $\sbullet$  \textbf{Div} (unary \& dyadic): Motion diversity. Temporal variance across a sequence of poses. Measures the amount of motion in a sequence.
    
    \noindent $\sbullet$ \textbf{Paired FD} (dyadic): The synchrony between  Alice and Bob's motion dynamics as a couple measured by the Frechet distribution distances on Alice-Bob \emph{pairs} (P-FD). Calculated FD on concatenated Alice and Bob pose (joint locations).

\bigskip
\noindent
\textbf{Baselines.}
We compare our dyadic prediction results to the following baselines:

    \noindent $\sbullet$  \textbf{Nearest Neighbor (NN) on Bob's input motion:} A segment-search method commonly used for synthesis in graphics. Given Bob's input motion, we find its nearest neighbor from the training set according to our learned VQ-VAE embedding vectors and use its corresponding ground truth Alice segment as the prediction.
    
    \noindent $\sbullet$   \textbf{Randomly-Chosen Training Alice (Random Train. Alice)}: Return a randomly-chosen Alice sequence from the training set.

    \noindent $\sbullet$  \textbf{Random Sample}:
    Random sample of the learned codebook indices. 
 
\subsection{VQ-VAE Codebook Ablation}

We experiment with different VQ-VAE codebook model variants by evaluating the effect on the reconstructed motion. The full results are presented in Table~\ref{tab:quantitative-vqvae}. More specifically, with ``Sep-Person'' we indicate that Alice and Bob have separate codebooks, while ``Joint-Person'' means that they share the same codebook.
Moreover, for ``Sep-Feat'' we learn three separate codebooks for the three factors of motion (body pose, orientation, translation), while for ``Joint-Feat'' we learn a unified codebook that entangles these three factors.
For the different variants we report the error in the motion reconstruction, as measured by the MPJPE, PA-MPJPE, and FD metrics.
We observe that learning three separate codebooks for body pose, orientation, translation leads to clear improvement in the reconstruction metrics which confirms our design choice.
Moreover, whether the codebooks for Alice and Bob are separate has smaller effect in the motion reconstruction, but we observed smaller standard deviation when we keep them separate, so this is the design we adopt.

\begin{table}[ht]
\centering
\setlength{\tabcolsep}{3pt}
\footnotesize
\begin{tabular}{@{}lccc@{}}
\toprule
VQ-VAE Model & MPJPE $\downarrow$ & PA-MPJPE $\downarrow$ & FD $\downarrow$ \\
\midrule

\rowcolor{Gray}
Sep-Person Sep-Feat & $0.112^{\pm 0.009}$ & $0.038^{\pm 0.000}$& $0.004^{\pm 0.001}$\\
Sep-Person Joint-Feat &	$0.272^{\pm 0.007}$&	$0.052^{\pm 0.001}$&	$0.014^{\pm 0.008}$\\

Joint-Person Sep-Feat & $0.112^{\pm 0.011}$& $0.038^{\pm 0.001}$& $0.006^{\pm 0.001}$\\
Joint-Person Joint-Feat & $0.261^{\pm 0.011}$& $0.051^{\pm 0.001}$& $0.116^{\pm 0.090}$\\
\bottomrule
\end{tabular}
\caption{\textbf{VQ-VAE Codebook Ablation.} Quantitative comparison of joint and separate VQ-VAE model variants in terms of MPJPE, PA-MPJPE, and FD metrics.}
\label{tab:quantitative-vqvae}
\end{table}

\subsection{Quantitative Results}
\label{sec:results}

\begin{figure*}
    \centering
    \includegraphics[width=0.9\linewidth]{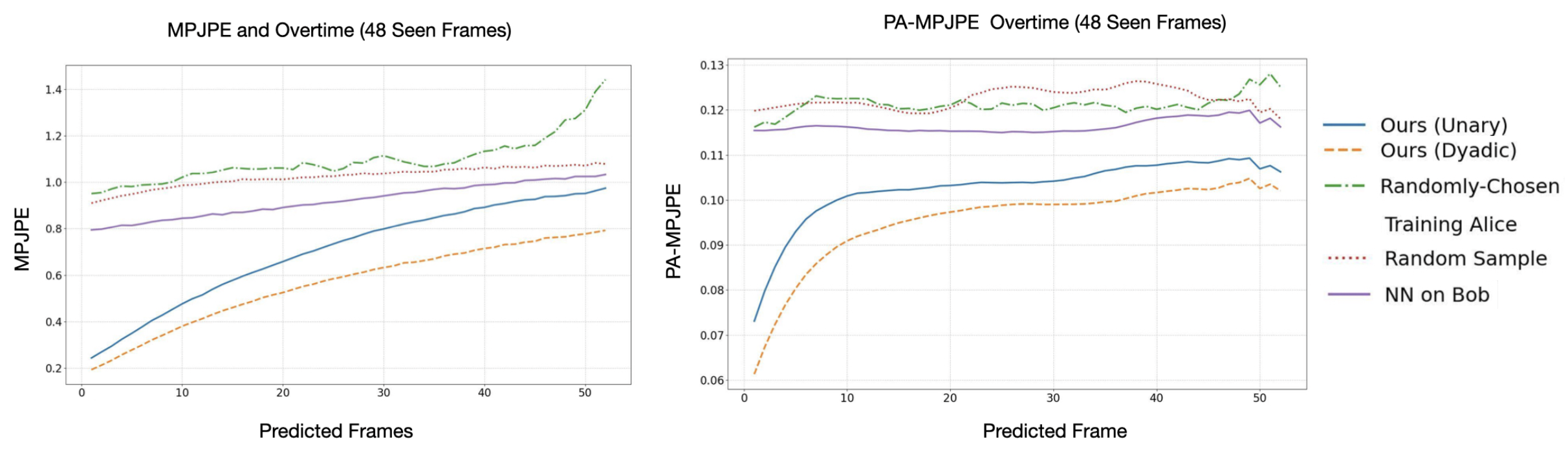}
    \vspace{-0.2cm}
    \caption{\textbf{Prediction error grows over time, more so for unary prediction.} Graphs of the MPJPE and PA-MPJPE metrics over time computed for ours vs.~baselines (in comparison with ground truth) starting from $t=t_\pi$, the point in which we start predicting. While our predicted future motion is correct initially in both conditions, prediction error grows over time faster in the unary than in the dyadic case. This increase is expected since motion during physical interaction highly depends on one's interaction partner. In contrast, all baselines start from a high error at $t_\pi$.}
    \label{fig:mpjpe-single-over-time}
    \vspace{-0.3cm}
\end{figure*}

We start by observing the temporal dynamics of our motion prediction method by graphing Alice's Mean Per Joint Position Error (MPJPE, PA-MPJPE) over time in Figure~\ref{fig:mpjpe-single-over-time}. MPJPE measures Euclidean distance between ground truth and predicted 3D joint locations averaged over 14 major body joints (no root alignment is performed). PA-MPJPE is the same loss after Procrustes Alignment on each frame, which ignores any translation, rotation or scale errors. 
Effectively, MPJPE jointly captures location and pose errors, while PA-MPJPE only captures local pose errors. 
We note that, as expected, while our predicted motion is accurate initially, it quickly digresses from the ground truth trajectory over time. The fact that this happens more rapidly for the unary versus the dyadic case suggests that our dyadic prediction successfully captures informative social signals from Alice's dance partner. We observe slower increase for PA-MPJPE because it only captures errors in the local pose, and unlike trajectories, which can diverge quickly, we rarely make gross errors in the predicted poses.

Our expectation that the additional information from Bob's motion should improve Alice's predicted motion is corroborated by our experiments as summarized in Table~\ref{tab:quantitative-concatenated}. As the table shows, the diversity (Div) of our predictions of Alice's motion is similar to the ground truth for both the unary and the dyadic cases. However, taking Bob's motion into account during dyadic prediction improves the realism (FD) of Alice's predicted motion compared to the unary case.

In the unary case, we note that while Figure~\ref{fig:mpjpe-single-over-time} demonstrated that the future predicted motion diverges from the specific single corresponding ground truth sequence, the overall realism (FD) of the predicted unary motion is similar to that of the baselines containing real motion sequences, NN on Bob and Random Train. Alice. However, in comparison, the realism (FD) of our dyadic prediction is much higher (\ie lower FD) since conditioning Alice's motion on that of her dance partner constrains the set of moves that she might perform and pushes the distribution over all her possible conditional motion trajectories closer to the ground truth Alice who dances with Bob.

Looking at the various baselines, we confirm that Random Sample is indeed the least realistic (FD), the most diverse (Div), and the least synchronous with Bob (P-FD). In comparison, choosing Alice randomly from the training data results in a realistic motion trajectory (FD) that is similar in diversity (Div) to the ground truth but is not synchronous with her partner, Bob (P-FD). Doing NN on Bob, outperforms the other baselines on all fronts, but it is still inferior compared to our predictions.

That said, Alice's motion is most synchronous with Bob (P-FD) when we condition her motion on Bob's in the dyadic case, compared to all the baselines, including NN on Bob, which looks for the nearest neighbor of Bob's motion trajectory and uses his ground truth partner as the prediction. This finding confirms our hypothesis that conditioning on an interaction partner's behavior not only improves motion realism and diversity of an individual but also enhances the synchrony of the couple as a whole.

\begin{table}\centering \footnotesize
\setlength{\tabcolsep}{3pt}
\begin{tabular}{@{}lrcrcr@{}}\toprule 
& \multicolumn{1}{c}{Realism} & \phantom{a} & \multicolumn{1}{c}{Diversity} & \phantom{a} & \multicolumn{1}{c}{Synchrony} \\
\cmidrule{2-2} \cmidrule{4-4} \cmidrule{6-6}
& {$\text{FD}$} $\downarrow$ && {$\text{Div}$} && {$\text{P-FD}$} $\downarrow$\\ 
\midrule
\rowcolor{light-gray}
\textit{GT} & - && \textit{0.694} && - \\

NN on Bob & 0.023 &&  0.773 && 0.079 \\
Random Train. Alice & 0.027 &&  0.791 && 0.107 \\
Random Sample & 0.212 &&  0.902 && 0.405 \\

\rowcolor{Gray}
Ours (unary) & 0.025 &&  0.697 && - \\
\rowcolor{Gray}
Ours (dyadic) & 0.012 &&  0.688 && 0.026 \\
\bottomrule
\end{tabular}
\caption{\textbf{Quantitative results.} Comparison against ground truth annotations (GT) on a held-out test set. $\downarrow$ indicates lower is better. For diversity, closer to GT is better.
}
\label{tab:quantitative-concatenated}
\end{table}

\subsection{Qualitative Results}
\begin{figure*}
    \centering
    \includegraphics[width=1\linewidth]{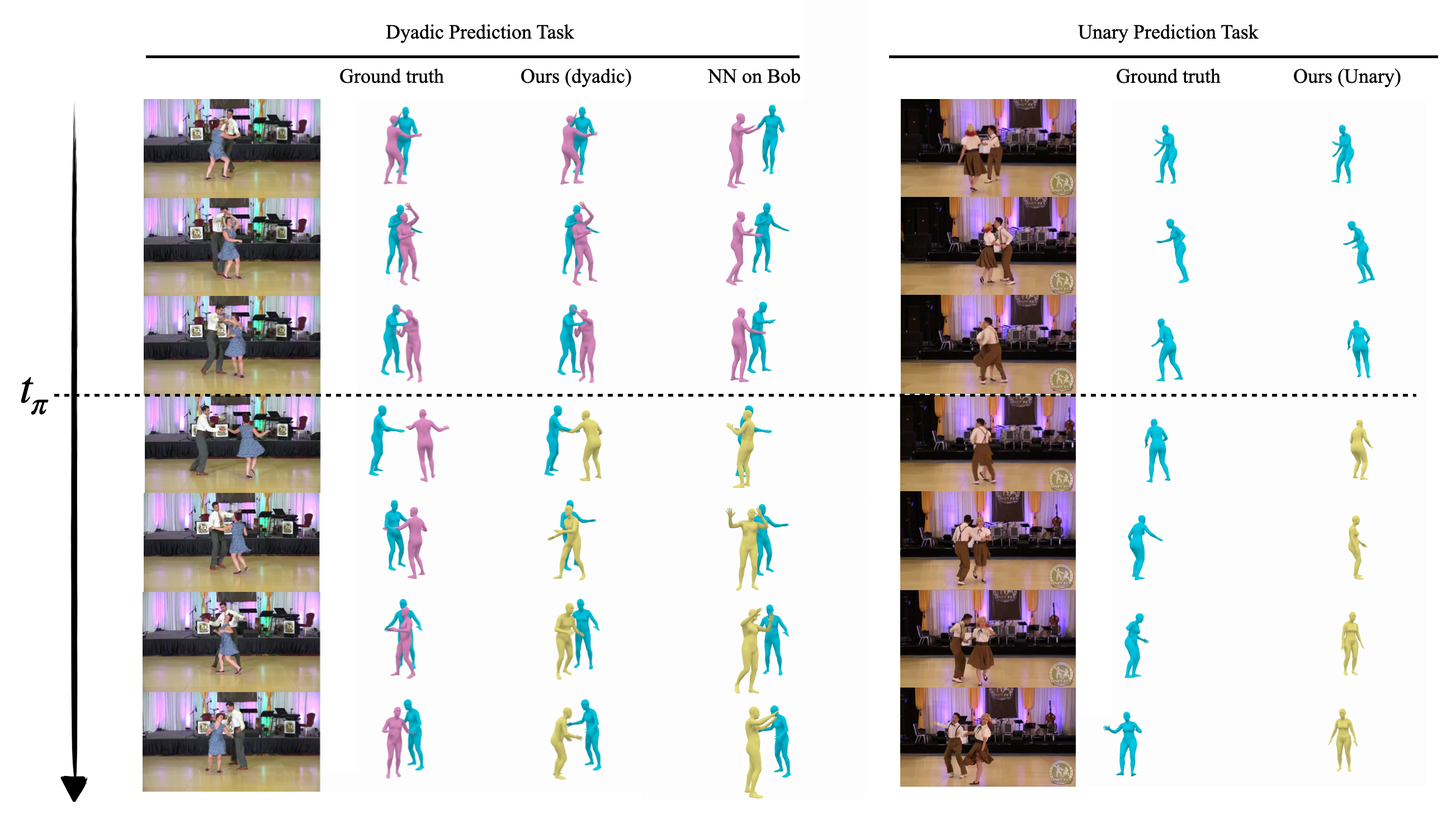}
    \vspace{-1.0cm}
    \caption{\textbf{Qualitative results of Alice's autoregressive motion prediction.} On the left, we input Alice and Bob's past motion up to frame $t_\pi$ and then start predicting Alice's future motion. We color the ``predicted'' Alices in yellow. On the right is the result of the unary prediction task. We input Alice's past motion (in blue) and predict her future motion (in yellow). Our model can predict plausible future motion of Alice including complex motions in couple dance like rotations.
    }
    \label{fig:qualitative}
\end{figure*}

We provide extensive qualitative results of our approach in the Supp. Video. Moreover, we present a few example results of our dyadic motion prediction in Figure~\ref{fig:qualitative}. We compare motion prediction in the dyadic and unary cases to the ground truth and NN on Bob in the dyadic case. While our dyadic prediction of Alice's motion results in a plausible pose trajectory that fits Bob's motion, the trajectory predicted by the NN baseline, while realistic-looking, does not match her partner.

\subsection{Limitations}

Unlike the majority of previous work, which relies on motion capture data that is often limited by heavy instrumentation and scalability issues, we study human motions recovered from Internet videos, extending the scope of our analysis. However, this approach means that the quality of our data heavily depends on the success of the video reconstruction method. Although we have utilized a state-of-the-art method~\cite{ye2023decoupling} for human motion recovery, we still encounter issues such as errors in the relative positions of the partners, potential mesh interpenetration, and imprecise contact estimation. Yet, a benefit of our approach is that it can leverage rapid advancements in 4D human motion reconstruction techniques. Combining the dyadic synchrony we have identified as crucial for motion prediction with these improvements presents an intriguing avenue for future work. 

\section{Conclusion}
This paper investigates the benefit of considering social information for behavior prediction during physical social interaction, such as couple dance. We demonstrate how to quantize full-body motion by disentangling motion into its constituent pose, orientation, and translation via a parametric body model. We train an autoregressive non-deterministic transformer predictor directly on the parameters of the parametric body model. We release a dataset of Swing dance in-the-wild videos with 3D pseudo ground truth on which we perform our experiments. Our results demonstrate that while we can predict a person's immediate future motion from past dance moves, considering the interaction partner provides a much richer context and results in more accurate predictions. While we focused on the dyadic scenarios in this work, our framework can be directly generalized and applied to multiple interaction partners in future work.

\paragraph{Acknowledgements} 
We thank Evonne Ng and members of the BAIR community for helpful discussions. This work was supported by BAIR/BDD sponsors, ONR MURI (N00014-21-1-2801), and the DARPA MCS program.

{
    \small
    \bibliographystyle{ieeenat_fullname}
    \bibliography{main}
}





\end{document}